\documentclass[12pt]{iopart}
\usepackage{graphicx}
\usepackage{iopams}
\usepackage[ruled,vlined]{algorithm2e}
\begin{document}
\title{Learning from Survey Propagation: a Neural Network for MAX-E-$3$-SAT}

\author{Raffaele Marino}

\address{Laboratoire de Th\'eorie des Communications,\\
  Facult\'e Informatique et Communications,\\
  \'Ecole Polytechnique F\'ed\'erale de Lausanne,\\
  1015, Lausanne, Switzerland\\}
\ead{raffaele.marino@epfl.ch}

\begin{abstract}
\small Many natural optimization problems are NP-hard, which implies that they are probably hard to solve exactly in the worst-case. However, it suffices to get reasonably good solutions for all (or even most) instances in practice. This paper presents a new algorithm for computing approximate solutions in ${\Theta(N})$ for the Maximum Exact 3-Satisfiability (MAX-E-$3$-SAT) problem by using deep learning methodology. This methodology allows us to create a learning algorithm able to fix Boolean variables by using local information obtained by the Survey Propagation algorithm. By performing an accurate analysis, on random CNF instances of the MAX-E-$3$-SAT with several Boolean variables, we show that this new algorithm, avoiding any decimation strategy, can build assignments better than a random one, even if the convergence of the messages is not found. Although this algorithm is not competitive with state-of-the-art Maximum Satisfiability (MAX-SAT) solvers, it can solve substantially larger and more complicated problems than it ever saw during training.
\end{abstract}


\submitto{\MST}
\maketitle

\section{Introduction}
\label{sec::intro}

The Boolean Satisfiability (SAT) Problem \cite{knuth2015art, cook1998combinatorial} is the issue of finding an assignment that satisfies a given Boolean formula. A Boolean formula is any operation made with Boolean variables, where each variable can take the value $TRUE$ or $FALSE$, $\{1, 0\}$ respectively. For example, a CNF (conjunctive normal form) \cite{whitesitt2012boolean} formula is a conjunction of one or more clauses, where a clause is a disjunction of literals. A CNF formula is satisfiable if and only if there is a configuration of the Boolean variables that simultaneously satisfy all the clauses.

In our work, $N$ defines the number of Boolean variables and $M$ the number of clauses so that the CNF formula has the following form:
\begin{equation}
\label{CNF:formula}
\bigwedge_{1\leq c \leq M}(\bigvee_{1\leq i \leq l_c} p_{ci}),
\end{equation}
where $l_c$ is the size of clause, i.e. the number of literals in clause $c$ for $1\leq c \leq M$, and $p_{ci}$ is a literal, thus a proposal variable $x_i$ or its negation $\overline x_i$, for $1\leq i \leq N$.

The maximization problem associated with SAT is called MAX-SAT. In this case, a solver tries to satisfy the maximum number of clauses given a CNF formula \cite{battiti1997reactive, lourencco2019iterated}. If a CNF formula has in each clause at least $k$ literals, then the problem is called MAX-$k$-SAT. If there are exactly $k$ literals for each clause in a CNF formula, then the problem is named MAX-E-$k$-SAT \cite{haastad2001some}.

The MAX-SAT is of considerable interest not only from the theoretical side but also for applications. For instance, many mathematical logic and artificial intelligence issues can be expressed in the form of satisfiability or some of its variants, like constraint satisfaction. Examples are in probabilistic inference \cite{walter2017constraint}, data analysis \cite{berg2019applications}, Maximum Clique and Maximum Independent Set\cite{san2019new, marino2018revisiting, marino2020large}, software analysis \cite{si2017maximum}, reasoning over bio networks and Bayesian network structure learning \cite{gouveia2020revision}, minimization of visibly pushdown automata \cite{heizmann2017minimization}, compressive sensing \cite{ayanzadeh2019sat}, community detection \cite{jabbour2020sat} and much more \cite{benedetti2018parametric, urbonas2020use}. For instance, in physics, the MAX-SAT problem is used for providing a provable periodically constrained ground state of a complex lattice \cite{huang2016finding}. Physicists also study the global landscape structure of the MAX-SAT problem for understanding phase transitions that appear into the solution space \cite{ochoa2020global}, trying to connect them to computational complexity limits. An interesting example of this research was developed in \cite{ercsey2011optimization, molnar2018continuous}. In these papers, continuous-time deterministic systems based on ordinary differential equations were proposed as SAT and MAX-SAT solvers. These works are based on the observation that the continuous-time deterministic systems have dynamics attracted by fixed points, identifying solutions with minimum energy.

From the theoretical point of view, the MAX-SAT problem is studied for giving optimal inapproximability results. Inapproximability results help to understand the computational complexity of hard problems \cite{haastad2001some, johnson1974approximation}. Many natural optimization problems, indeed, are NP-hard. This implies that they are probably hard to solve exactly in the worst-case. The worst-case complexity measures the maximum amount of resources that an algorithm requires, given an input of arbitrary size \cite{cormen2009introduction}.

However, it suffices to get reasonably good solutions for all (or even most) instances in practice. Examples of these results have been studied since 1973 when Johnson \cite{johnson1974approximation} analyzed the worst-case behavior of simple, polynomial-time, random algorithms for finding approximate solutions to various combinatorial optimization problems. He measured the worst solution value ratio, which can be reached by an algorithm, to the optimal one. For a maximization problem, an algorithm is a $\rho$-approximation algorithm, $\rho < 1$, if it produces a solution whose objective value is at least $\rho \cdot OPT$ where $OPT$ is the global optimum, for each instance. A similar definition applies to minimization problems. The approximation algorithms' important property relates the size of the solution produced directly to a lower bound on the optimal solution. Instead of telling us how well we might do, they will tell us about the worst-case, i.e., how badly we might perform \cite{skiena1998algorithm}.

Following this research topic, many computer scientists have proven rigorous results over optimal inapproximability. The first result proving hardness for the problem we are discussing here was obtained in the fundamental paper by Arora et al. \cite{arora1998proof}. He established the PCP theorem. The theorem states that every decision problem in the NP complexity class has probabilistically checkable proofs, where a verifier reads only a constant number of bits and uses logarithmic random of bits. Significant results were obtained successively by Bellare, Sudan, and others in \cite{bellare1998free, bellare1996linearity, poloczek2017greedy, chou2020optimal, brakensiek2021mysteries}, but the most famous was given by H{\aa}stad in 1997 \cite{haastad2001some}. He proved optimal inapproximability results, up to an arbitrary $\epsilon>0$, for MAX-E-$k$-SAT for $k \geq 3$, by maximizing the number of satisfied linear equations in an over-determined system of linear equations modulo a prime $p$. More precisely, the author showed that for the MAX-E-$3$-SAT, no approximate algorithm could outperform the random assignment threshold, which is set to be $7/8$ the optimal one, unless $P=NP$. He also stated that the MAX-E-$4$-SAT is not approximable beyond the random assignment threshold on satisfiable instances. The random assignment threshold is set to be $15/16$ the optimal one, unless $P=NP$. These results are only valid for approximate algorithms.

In contrast to this kind of algorithms, heuristics \cite{schneider2007stochastic} can find better approximate solutions. However, their worst-case performance can be tough to analyze and, therefore, heuristic methods may be considered approximate and not accurate algorithms. Although they have this negative reputation, heuristics are the only viable option for various optimization problems that need to be routinely solved in real-world applications.  

Heuristic algorithms for solving MAX-SAT problems can be roughly classified into two main categories. The first category of algorithms searches for a solution by performing a biased random walk in the space of configurations. Instead, the second one tries to build a solution assigning variables, according to some estimated marginals. MaxWalkSAT, focused Metropolis search or diffusion Monte Carlo algorithms, genetic algorithms with local search, and many other SAT-Solvers belong to the former category \cite{selman1993local, bouhmala2016walksat, kirkpatrick1983optimization, liu2017should, bouhmala2019combining, djenouri2017data, brandts2019smoothed, jarret2016adiabatic, traversa2018evidence, ali2019solving, bouhmala2018combining, berend2020effect, xu2019iterative}. In contrast, in the second category, we find algorithms deriving from the class of message passing algorithms. Examples are Warning Propagation (WP), Belief Propagation Guided Decimation (BPGD), Survey Inspired Decimation (SID), Backtracking Survey Propagation (BSP), and SP-$y$, a generalization of Survey Propagation (SP) algorithm \cite{montanari2007solving, mezard2002analytic, braunstein2005survey, marino2016backtracking, battaglia2004minimizing, chieu2009relaxed, wang2017warning}. The last algorithms use a decimation move for building a solution to the problem. This decimation procedure modifies the underlying graph's structure and makes these algorithms hard to be analytically analyzable.

This paper introduces a new heuristic-learning algorithm that collects local information from the Survey Propagation algorithm, avoids any decimation strategy, and fixes the local variables by using deep learning methodology \cite{lecun2015deep, goodfellow2016deep, krizhevsky2012imagenet}. 

The deep learning methodology is part of a broader family of machine learning methods based on artificial neural networks. It has been applied in many fields, from computer science to physics and economics, with many useful applications. Deep learning uses multiple layers to extract higher-level features from the raw input. Here, the deep learning methodology extracts patterns from instances of MAX-E-$3$-SAT and their solutions for learning how to fix a single Boolean variable.
 
In combinatorial optimization, deep learning methodology has been used for solving SAT problems or problems related to graphical models. For instance,  Selsam et al. \cite{selsam2018learning} obtained exciting results for SAT problems. More precisely, the authors present a message passing neural network that learns to solve SAT problems after only being trained as a classifier to predict satisfiability. Instead, for problems related to graphical models, Dai et al.\cite{Dai2019} present a learning algorithm for graph problems using a unique combination of reinforcement learning and graph embedding. Many other results have been reached during this period, and we refer to an interesting survey on machine learning for combinatorial optimization of Bengio et al. \cite{bengio2020machine} and references therein. Regarding the MAX-SAT Problem, instead, an exciting work based on statistical learning theory has been presented recently in \cite{kumar2020learning}. They introduce a novel setting for learning combinatorial optimization problems from contextual examples.  

These works deal with small instances because the neural networks are fed with the whole instance of the problem. In this paper, in contrast, we present, for the first time, as far as we know, a linear algorithm that takes information locally on the graph and uses a deep neural network for assigning a single variable. This strategy allows us to analyze CNF formulae composed of $10^6$ variables, larger and more difficult than the neural network ever saw during training. 

The motivation that guides us in simplifying heuristic methods is the following: although making a heuristic more complicated does not necessarily make it better in the worst-case, maybe making it simpler does not necessarily make it worse in the worst-case. The worst-case scenario for heuristic message passing algorithms is due to the fact that a full convergence of the messages is not found. This can appear even for a single message. When such a scenario appears, we cannot say anything about the instance we are looking at. We lose all the local information that is correctly obtained by the message passing procedure.

For this reason, we deal with SP equations. Although they are not the best for the MAX-SAT problem, in contrast to the SP-$y$ equations, they are simpler and much more suitable for meeting the worst-case scenario. They may make this heuristic-learning algorithm analytically analyzable. Indeed, decimation and backtracking moves are avoided.

This paper is divided into the following sections: the first one recalls the MAX-E-$3$-SAT problem and its factor graph representation; the second one recalls the Survey Propagation algorithm; the third one introduces the deep neural network and also presents the numerical analysis. We conclude our manuscript with a discussion on the future research directions that this new method gives rise to.

\section{The MAX-E-3-SAT Problem and its Factor Graph Representation}

\begin{figure}
  \centering
  \includegraphics[width=0.45\columnwidth]{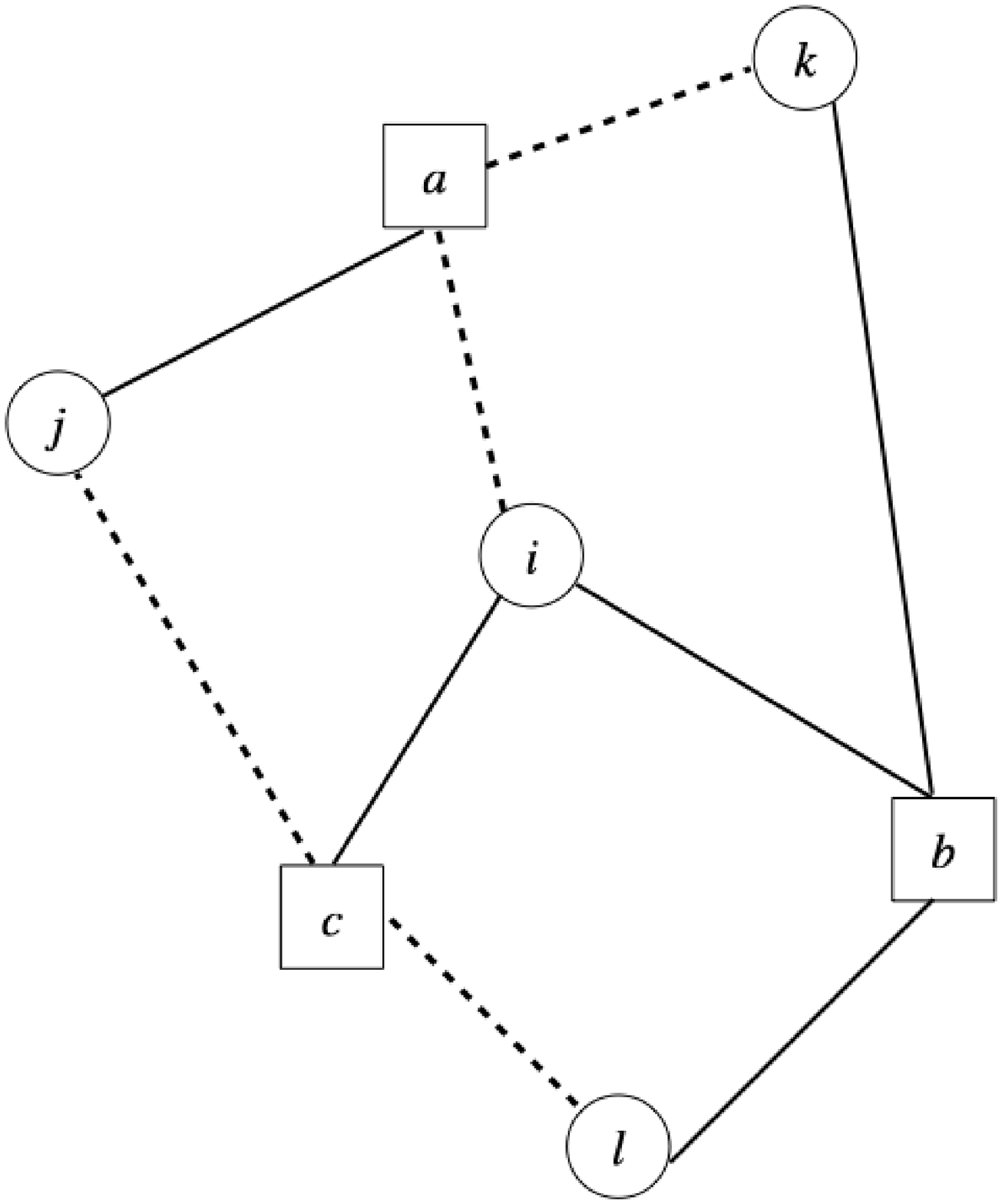}
  \caption{\small The figure shows the representation as a factor graph of a $3$-SAT (MAX-E-$3$-SAT) problem. Squares are functional nodes, while circles are variable nodes. A functional node is connected with an edge to a variable node if and only if the variable appears as a literal into the clause. If the edge is a dashed line, then the literal is negated into the clause. Otherwise, if the edge is a full line, the literal appears not negated. } 
  \label{fig:factgaraph}
\end{figure}

As explained in the \Sref{sec::intro}, the MAX-SAT is the maximization problem associated with the SAT problem. It is asked to find an assignment of the Boolean variables such that the maximum number of clauses, in a CNF formula, is satisfied. The maximization version of the $k$-SAT problem is called MAX-E-$k$-SAT, in this case, each clause contains exactly $k$ literals. With $k=3$ a $3$-SAT problem with $N=9$ variables and $M=4$ clauses is of the form:
\begin{equation}
\label{3-sat}
(x_1 \vee x_2 \vee x_3) \wedge (\overline{x}_1 \vee x_4 \vee x_5) \wedge (\overline{x}_2 \vee x_6 \vee x_7) \wedge (\overline{x}_3 \vee x_8 \vee x_9).
\end{equation}

Clearly an assignment that satisfies all the clauses is Sol$_{3-SAT}:(x_1=1, x_2=0, x_3=0, x_4=1, x_5=0, x_6=1, x_7=0, x_8=1, x_9=0)$. The MAX-E-$3$-SAT looks for an approximate solution of the problem in \eref{3-sat}. For example, a solution of the MAX-E-$3$-SAT, outputted by a random algorithm, could be Sol$_{MAX-E-3-SAT}:(x_1=0, x_2=0, x_3=0, x_4=1, x_5=0, x_6=1, x_7=0, x_8=1, x_9=0)$. Sol$_{MAX-E-3-SAT}$ does not satisfy all the clauses in \eref{3-sat}, but it is an approximate solution where just a clause is unsatisfied, i.e. $(x_1 \vee x_2 \vee x_3)=FALSE$. For a general instance of the MAX-E-$3$-SAT problem that contains $N$ variables and $M$ clauses, it is easy to see that a random assignment satisfies each clause with probability $7/8$. Hence, if there are $M$ clauses, it is not hard to find an assignment that satisfies $7M/8$ clauses. Since we can never satisfy more than all the clauses this gives a $7/8$-approximation algorithm \cite{johnson1974approximation, haastad2001some}. 
In this paper, as stated in the \Sref{sec::intro}, we are interested in heuristic methods that use message passing procedure. For this reason, we recall the factor graph representation of satisfiability problems. The SAT problem, and thus the MAX-SAT, is represented as a factor graph where clauses are identified as functional nodes and variables as variable nodes. A factor graph is a bipartite graph representing the factorization of a function.

In \Fref{fig:factgaraph}, it is shown a cartoon of the factor graph associated with a $3$-SAT (MAX-E-$3$-SAT) problem. The variables, $i, j, l$ and $k$, are variable nodes (circles) while clauses, $a,b$ and $ c$, are functional nodes (squares). Each variable enters a clause as a literal, e.g., $p_{ci}$, if and only if it is connected with an edge. Dashed edges identify literals where variables are negated, while full edges identify literals where variables are not negated.
 
With the symbol $\partial_a$, we define the set of variables nodes that are connected with the functional node $a$, i.e., the literals of clause $a$. In contrast, with the symbol $\partial_i$, we define the set of functional nodes connected with the variable node $i$, i.e., the set of clauses where the literal indexed $i$ appears. The cardinality of the set $\partial_i$ is the degree of a variable node $i$, i.e., the number of links connected to a circle, and is defined with $n_i$. The set $\partial_i$ is also composed of two other sets, namely $\partial_i^+$ that contains the functional nodes where the variable node $i$ appears not negated, and $\partial_i^-$ that contains the functional nodes where the variable node $i$ appears negated. Obviously, the relation $\partial_i= \partial_i^+ \cup \partial_i^-$ holds. We also defined two more quantities, $n_i^{\pm}$, for the number of dashed and full edges of a variable node $i$. More precisely, $n_i^{+}$ defines the cardinality of the set $\partial_i^+$, while $n_i^{-}$ defines the cardinality of the set $\partial_i^-$.

With the symbol $\partial_{ia}^{+}$ (respectively $\partial_{ia}^{-}$) we define the set of functional nodes containing the variable node $i$, excluding the functional node $a$ itself, satisfied (respectively not satisfied) when the variable $x_i$ is assigned to satisfy clause $a$. In other words, if the variable $x_i$ is not negated in the clause $a$, then the $\partial_{ia}^{+}$ is the set of functional nodes containing the variable node $i$, excluding the functional node $a$ itself, where the variable node $i$ is connected with a full edge, thus where the variable $x_i$ appears not negated, while $\partial_{ia}^{-}$ is the set of functional nodes containing the variable node $i$, where the variable node is connected with dashed edges, thus where the variable $\overline{x}_i$ appears negated. In contrast, if the variable $\overline x_i$ is negated in the clause $a$, then the $\partial_{ia}^{+}$ is the set of functional nodes containing the variable node $i$, excluding the functional node $a$ itself, where the variable node $i$ is connected with a dashed edge, thus where the variable $\overline{x}_i$ appears negated, while $\partial_{ia}^{-}$ is the set of functional nodes containing the variable node $i$, where the variable node is connected with full edges, thus where the variable $x_i$ appears not negated.

\section{The Survey Propagation Algorithm}
\label{sec::SPalgo}

The Survey Propagation algorithm (SP) is a heuristic message passing algorithm. A detailed description of the Survey Propagation algorithm can be found in \cite{mezard2002analytic, braunstein2005survey, marino2016backtracking, mezard2009information}, here we recall it naively. Mezard, Parisi, and Zecchina developed it in \cite{mezard2002analytic} from the assumption of one-step replica symmetry breaking and the cavity method of spin glasses. SP has been applied to different combinatorial optimization problems, like random K-SAT, MAX-E-$k$-SAT, q-coloring, Maximum Independent Set, etc. \cite{mezard2002analytic, braunstein2005survey, marino2016backtracking, battaglia2004minimizing, chieu2009relaxed, mulet2002coloring, barbier2013hard}, always showing the best performance for solving these problems. It works on a factor graph underlying the CNF formula. For $N \to \infty$, SP is conjectured to work better and better because it runs over locally-tree like factor graphs, and cycles into the graph are at least $\Or(\log N)$.

Broadly speaking, SP exchanges messages between variables and clauses for guessing the value that each variable needs to be set. More precisely, a message of SP, called a survey, passed from one function node $a$ to a variable node $i$ (connected by an edge) is a real number $\eta_{a\to i} \in [0, 1]$. Under the assumption that SP runs over a tree-like factor graph, the messages have a full probabilistic interpretation. In particular, the message $\eta_{a\to i}$ corresponds to the probability that the clause $a$ sends a warning to variable $i$, telling which value the variable $i$ should adopt to satisfy itself \cite{Maneva2007, braunstein2005survey}.

The updating rules of a single message $\eta_{a\to i} $ are presented in Algorithm $1$. 

 \begin{minipage}{0.85\textwidth}
 \begin{algorithm}[H]
 \label{algoSP}
\SetAlgoLined
\KwIn{set of all messages arriving onto each variable node $j\in \partial_a \setminus i$ }
\KwOut{new value for the message $\eta_{a\to i} $.}
 \For{$j\in \partial_a \setminus i$}{
 $s_{j\to a}^{-}=\left[1- \prod_{b \in \partial_{ja}^{-}} (1-\eta_{b\to j}) \right]\prod_{b \in \partial_{ja}^{+}} (1-\eta_{b\to j})$\;
 $s_{j\to a}^{+}=\left[1- \prod_{b \in \partial_{ja}^{+}} (1-\eta_{b\to j}) \right]\prod_{b \in \partial_{ja}^{-}} (1-\eta_{b\to j})]$\;
 $s_{j\to a}^{0}=\left[\prod_{b \in \partial_{j} \setminus a} (1-\eta_{b\to j}) \right]$\;
 if a set $\partial_{ja}^{\pm}$ is empty, the corresponding product takes value $1$ by definition\;
 }
 \Return $\eta_{a\to i}=\prod_{j \in \partial_{a} \setminus i} \left[ \frac{s_{j\to a}^{-}}{s_{j\to a}^{-}+s_{j\to a}^{+}+s_{j\to a}^{0}}\right]$
 \caption{Subroutine SP-UPDATE( $\eta_{a\to i} $)}
\end{algorithm}
 \end{minipage}
 
SP is a local algorithm that extracts information on the underlying graph of a CNF formula. As Input, it takes a CNF formula of a Boolean Satisfiability Problem, and it performs a message passing procedure to obtain convergence of the messages. More precisely, we are given a random initialization of all messages, and at each iteration, each message is updated following the SP-UPDATE rule described in Algorithm $1$. SP runs until all messages would satisfy a convergence criterion. This convergence criterion is defined as a small number $\epsilon$ such that the iteration is halted at the first time $t^*$ when no message has changed by more than $\epsilon$ over the last iteration. If this convergence criterion is not satisfied after $t_{max}$ iterations, SP stops and returns a failure output. Once a convergence of all messages $\eta_{a \to i}$ is found, SP's goal is to minimize the number of violated clauses. For doing that, a new strategy for fixing the value of the variables must be introduced. This strategy is called decimation and transforms the SP into the Survey Inspired Decimation Algorithm (SID). For using decimation, however, one needs to compute the SP marginals for each variable $i$:
\begin{equation}
\label{SID}
\eqalign 
S_i^-=\frac{\pi_i^-(1-\pi_i^+)}{1-\pi_i^+\pi_i^-},\\
S_i^+=\frac{\pi_i^+(1-\pi_i^-)}{1-\pi_i^+\pi_i^-},\\
S_i^0=1-S_i^--S_i^+,
\end{equation}  
where:
\begin{equation}
\pi_i^{\pm}=1-\prod_{b\in \partial_i^{\pm}}(1-\eta_{b \to i}).
\end{equation}

The SP marginal $S_i^{+}$ ($S_i^{-}$) tells the probability that the variable $i$ must be forced to take the value $x_i=1$($x_i=0$), conditional on the fact that it does not receive a contradictory message, while $S_i^{0}$ provides the information that the variable $i$ is not forced to take a particular value. 

Once all the SP marginals have been computed, the decimation strategy can be applied. Decimating a variable node $i$ means fixing the variable to $TRUE$ or $FALSE$ depending on the SP marginals, removing all satisfied functional nodes and the variable node $i$ from the factor graph, and removing all the literals into the clauses that have not been satisfied by the fixing. However, how to choose the variable node $i$ to decimate? The answer is simple, just selecting a variable with the maximum bias $S_{C_i}=1-\min(S_i^-, S_i^+)$. Decimated the variable node $i$, the SID iteratively runs the SP algorithm and uses decimation again. The decimation procedure continues till one of these three different outcomes appears: (i) a contradiction is found, then SID returns exit failure; (ii) SP does not find a convergence, then SID returns exit failure; (iii) all the messages converge to a trivial fixed point, i.e., all the messages are equal to $0$, in this case, SID calls WalkSAT, which solves the residual formula and builds the complete solution of the problem.

The SID has extremely low complexity. Each SP iteration requires $\Or(N)$ operations, which yields $\Or(N t_{max})$, where $t_{max}$ is the maximum time allowed for finding a convergence, i.e. a big constant. In the implementation described above, the SID has a computational complexity of $\Or(t_{max} N^2 \log N )$, where the $N\log N$ comes from the sorting of the biases. This can be reduced to $\Or(Nt_{max}(\log N)^2)$ by noticing that fixing a single variable does not affect the SP messages significantly. Consequently, SP can be called every $N\delta$ decimation step by fixing a fraction of variables at each decimation step. The efficiency of SID can be improved by introducing a backtracking strategy or a reinforcement strategy. We refer to \cite{chavas2005survey, parisi2003backtracking, marino2016backtracking, montanari2011reconstruction}, and references therein for a complete explanation of these strategies. 

\section{The Neural Network and a new heuristic-learining algorithm}

This manuscript aims to present a new heuristic-learning algorithm that can find an assignment for a set of Boolean variables that maximizes the number of satisfied clauses of a given CNF formula. Although this new heuristic-learning algorithm does not reach state-of-the-art algorithms for the MAX-SAT problem, it can solve problems that are substantially larger and more difficult than it ever saw during training. 
The code was developed in C++ using mlpack, a fast and flexible C++ machine learning library \cite{mlpackpaper}. The experiments were performed on a cluster with 128 cores and 512 GB of RAM. The code, the training data, and the test data can be downloaded from \cite{Marino2019GITHUB}.

\subsection{Empirical analysis of SP equations}

\begin{figure}
 \centering
 \includegraphics[width=0.45\columnwidth]{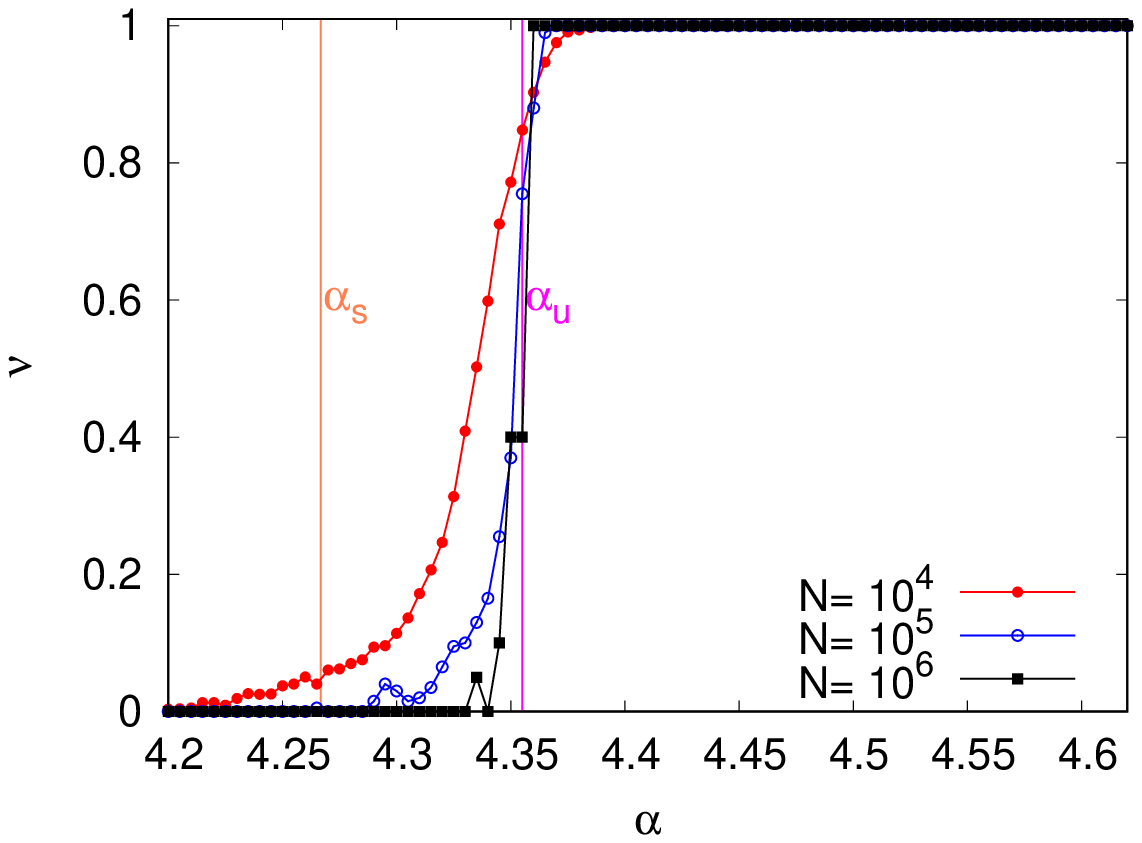}
 \includegraphics[width=0.45\columnwidth]{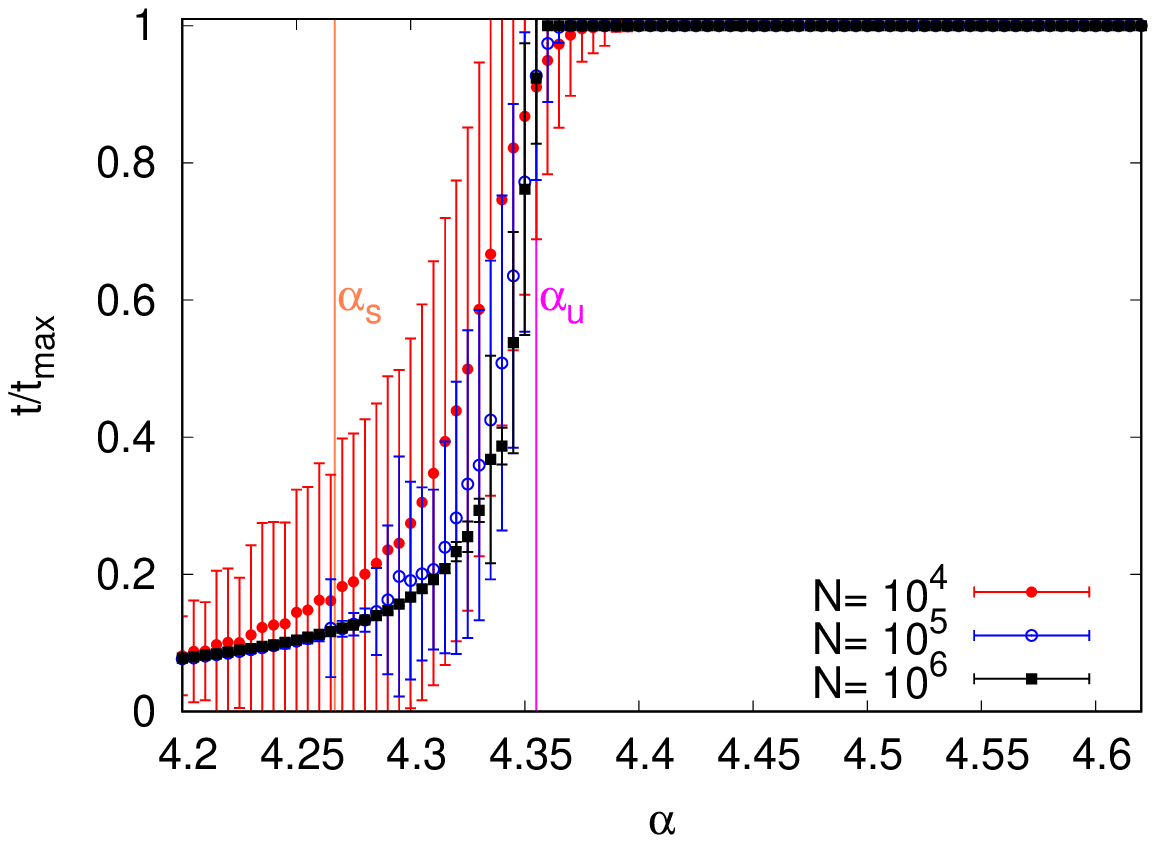}
  \caption{\textbf{Left}: The plot displays the fraction of random $3$-SAT 
 instances that did not converge, i.e. $\nu$, as a function of the clause density $\alpha$. 
  \textbf{Right}: The plot displays the average number of iterations $t$ needed by SP for finding a convergence for all messages, with $\epsilon=0.01$, normalized to $t_{max}=1024$ as a function of the clause density $\alpha$. In both cases, we analized $10^3$ instances for $N=10^4$ (red circle points), $10^2$ instances for $N=10^5$ (blue empty circle points), and $10$ instances for $N=10^6$ (black square points). The vertical coral line identifies the SAT-UNSAT threshold at $\alpha_s=4.267$. The vertical magenta line identifies the SP unique solution threshold  at $\alpha_u=4.355$. Error bars are standard deviations.}
\label{fig::frac_conv}
\end{figure}

\begin{figure}
 \centering
 \includegraphics[width=0.45\columnwidth]{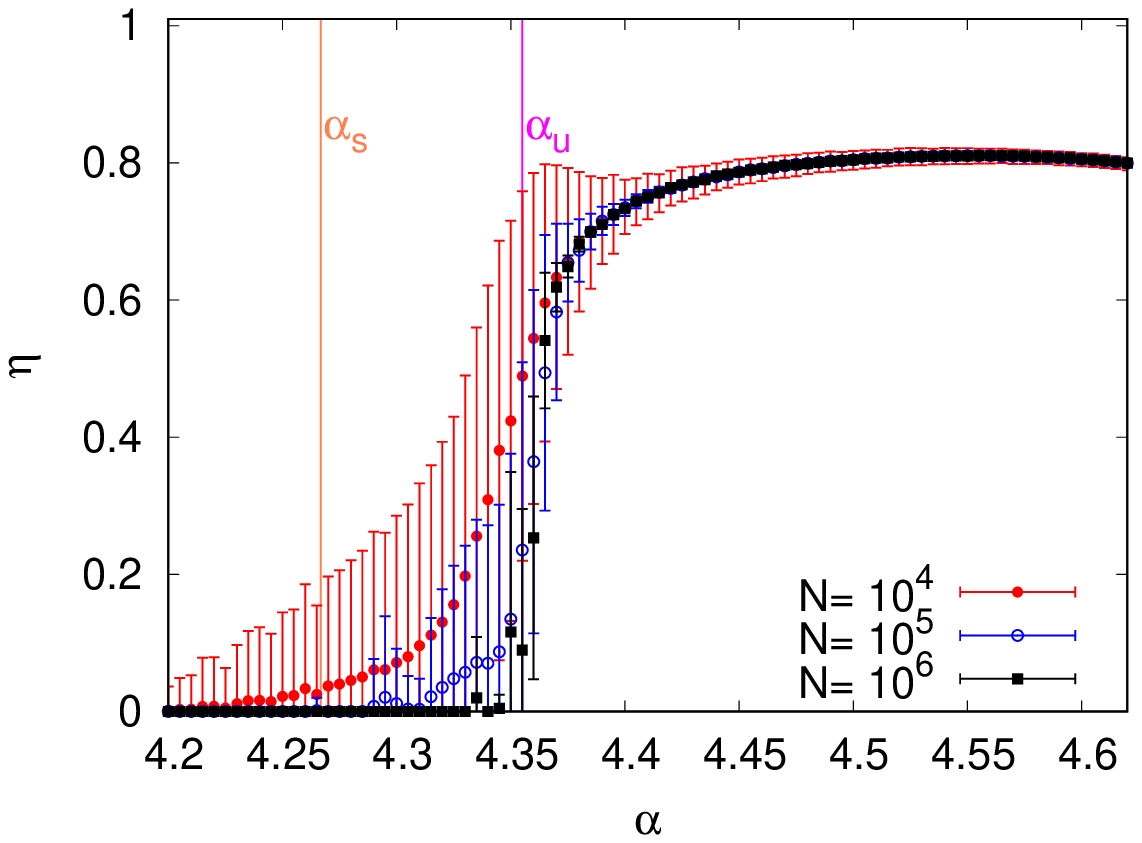}
  \includegraphics[width=0.45\columnwidth]{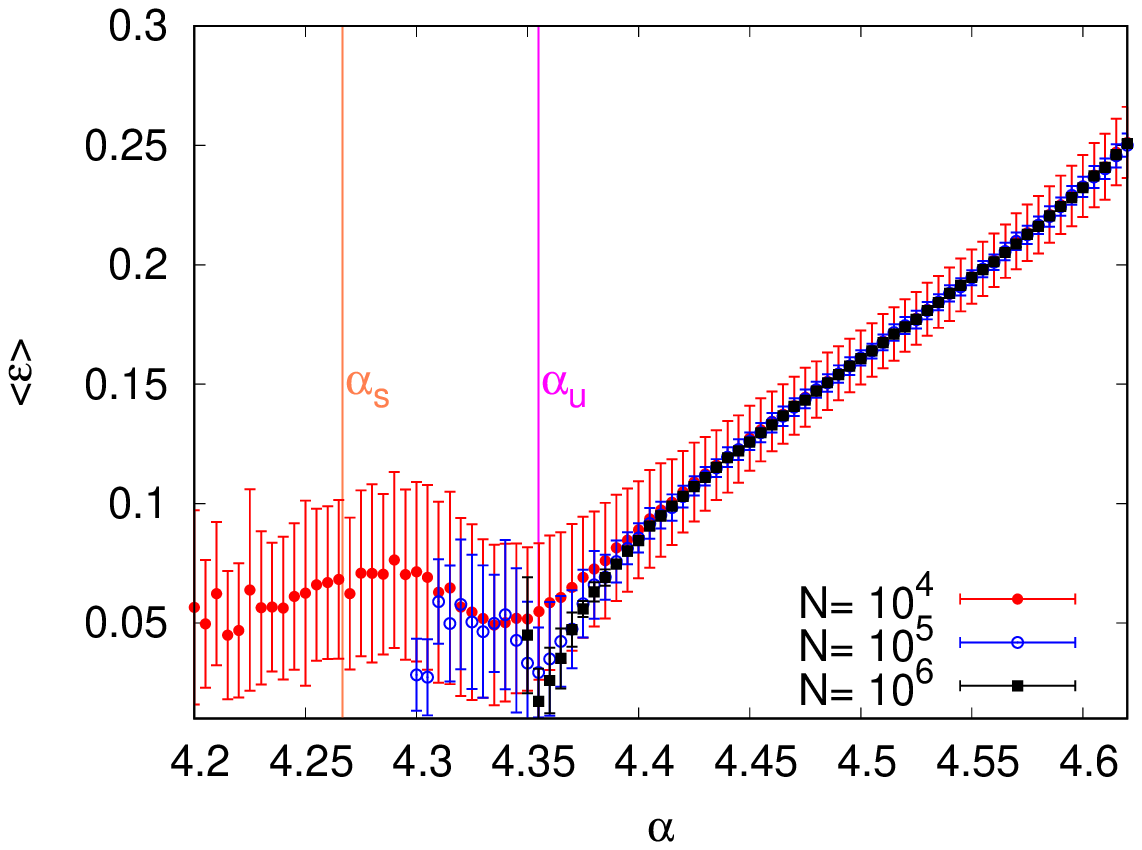}
  \caption{\textbf{Left}: The plot displays the average fraction of messages that do not converge in each instance of a random $3$-SAT, $\eta$, as a function of $\alpha$. For the random $3$-SAT problem, we analyzed $10^3$ instances for $N=10^4$ (red circle points), $10^2$ instances for $N=10^5$ (blue empty circle points), and $10$ instances for $N=10^6$ (black square points). The vertical coral line identifies the SAT-UNSAT threshold at $\alpha_s=4.267$. The vertical magenta line identifies the SP unique solution threshold at $\alpha_u=4.355$. Error bars are standard deviations. \textbf{Right}: The plot displays the average error $\langle \epsilon \rangle$ that the not converging messages commit at the last iteration $t_{max}$ as a function of the clause density $\alpha$, on only instances that do not find a full convergence of the messages. Before the $\alpha^{conv}_{3-SAT}$ the average error $\langle \epsilon \rangle$ seems to be bounded, i.e., $\langle \epsilon \rangle \leq 0.15$, while beyond the threshold, it grows linearly with $\alpha$, showing that more difficult is an instance because more clauses are present there, less accurate information can be obtained locally from the message passing procedure. For the random $3$-SAT problem, we analyzed $2\,10^3$ instances for $N=10^4$ (red circle points), $10^2$ instances for $N=10^5$ (blue empty circle points), and $10$ instances for $N=10^6$ (black square points).The vertical coral line identifies the SAT-UNSAT threshold at $\alpha_s=4.267$. The vertical magenta line identifies the SP unique solution threshold at $\alpha_u=4.355$. Error bars are standard deviations.}
  \label{fig::mess_conv}
\end{figure}

Before presenting the whole algorithm, we start to analyze the SP algorithm. It is known that the SP algorithm collects information locally and by using equations in \eref{SID} predicts the marginal probabilities that allow fixing a Boolean variable. 
This information, however, can be achieved only by a full convergence of the messages. Without a full convergence of the messages, algorithms based on SP equations always return a failure output. Therefore, for understanding the limits of SP, we perform an accurate analysis on a set of random $3$-SAT instances for different values of the clause density (the ratio of the number of clauses to the number of variables) $\alpha=M/N$, with large $M$, large $N$, and keeping constant $\alpha$. Instances are generated by considering $N$ variables and $M=\alpha N$ clauses, where each clause contains exactly $k=3$ distinct variables, and is picked up with uniform probability distribution from the set of $N!(k!(N-k)!)^{-1}$ $2^k$ possible clauses \cite{selman1993local, braunstein2005survey}.

We choose random $3$-SAT instances for two reasons. The first one is just for the sake of simplicity. We use the same instances to analyze the new algorithm's performance in approximating the solutions of the MAX-E-$3$-SAT problem associated with them. The second one, instead, is given by the fact that many theoretical results are well known. For example, it is known that the SAT-UNSAT threshold for the random $3$-SAT is at $\alpha_s=4.267$ \cite{ding2015proof, mertens2006threshold, bartha2019breaking}. The SAT-UNSAT threshold defines two regions sharply: for $N\to \infty$, the region before the threshold contains all the instances of random $k$-SAT problems that have at least an assignment that satisfies all the clauses (SAT region), while beyond the threshold, no assignment that satisfies all the clauses exists (UNSAT region). It is also known that when $\alpha=\alpha_U\approx 4.36$, the SP equations do not have a unique solution. This fact is not of direct importance for the random $3$-SAT problem because we are beyond the SAT-UNSAT threshold. No exact solution exists, i.e., not all the clauses of an instance can be satisfied simultaneously. However, for the MAX-E-$3$-SAT problem, this point is interesting. Indeed, from there, we expect that SP equations will not converge, and therefore the worst-case scenario for the SP algorithm appears.

We start analyzing the empirical convergence of the SP algorithm (initialized with uniformly random messages) as a function of the clause density $\alpha$, for different values of $N$. We fix, as described in the previous section, the value of $t_{max}$ to $1024$ and $\epsilon$ to $10^{-2}$. 

In \Fref{fig::frac_conv}, we plot, for random $3$-SAT, the fraction of instances that did not converge, $\nu$, (left panel), and the number of iterations $t$ that SP needs to make for reaching a convergence of all messages (right panel), as a function of the clause density $\alpha$. The analysis shows that the SP algorithm returns a failure output for random $3$-SAT at $\alpha_{3-SAT}^{conv}=4.355$ because a full convergence is not found, in agreement with the results obtained by Mezard and Montanari in \cite{mezard2009information}. In both cases, we observe a step function form of the fraction of instances that did not converge for $N \to \infty$. For $N \to \infty$, therefore, SP always converges before the specific value of the clause density $\alpha^{conv}_{3-SAT}$, because the solutions of the SP equations are unique, while does not converge beyond the $\alpha^{conv}_{3-SAT}$, because the SP equations have many solutions \cite{parisi2003probabilistic}.  This property shows that no algorithm, based on SP equations presented in Algorithm \ref{algoSP}, can build any solution beyond $\alpha_U=\alpha^{conv}_{3-SAT}$. Analyzing the average fraction of messages that do not converge, $\eta$, as a function of $\alpha$, however, it seems that beyond the convergence threshold $\alpha^{conv}_{3-SAT}$, a fraction of messages always converges. More precisely, a fraction of converging messages, which is almost $\sim20\%$ of the messages in each random $3$-SAT instance (see \Fref{fig::mess_conv}, left panel), exists. This fact inspires us to look at the average error of convergence. We define the average error of convergence as the quantity $\langle \epsilon \rangle$ such that:
\begin{equation}
\label{averageerror}
\langle \epsilon \rangle=\frac{1}{\mathcal{M}}\sum_{s=1}^{\mathcal{M}} \frac{1}{kM-n^{conv}_s} \sum_{i=1}^N \sum_{b\in{\partial_i}} \Delta_{b\to i}(t_{max}),
\end{equation}
where $k=3$, because we have $3$ messages for each clause, $n^{conv}_s$ is the number of messages that converge in the $s$-th instance over the $\mathcal{M}$ analyzed, and
\begin{equation}
\Delta_{b\to i}(t_{max})=\cases{|\eta^{t_{max}}_{b \to i}-\eta^{t_{max}-1}_{b \to i}|&for $|\eta^{t_{max}}_{b \to i}-\eta^{t_{max}-1}_{b \to i}| \geq \epsilon$\\
0&for $|\eta^{t_{max}}_{b \to i}-\eta^{t_{max}-1}_{b \to i}| < \epsilon$\\}
\end{equation}

In the case where $n^{conv}_s=kM$, we define that $\frac{1}{kM-n^{conv}_s} \sum_{i=1}^N \sum_{b\in{\partial_i}} \Delta_{b\to i}(t_{max})\,=0$.

In the right panel of \Fref{fig::mess_conv}, we plot the quantity $\langle \epsilon \rangle$ as a function of the clause density $\alpha$. For $N \to \infty$, the probability of finding a full convergence in the region $\alpha \in [4.200, 4.355)$ is equal to one because SP equations run over a factor graph that is locally tree-like. When $N$ is finite and small, for instance, $N=10^4$, the property of having a tree-like structure is not always preserved, and, therefore, we can meet before the threshold $\alpha^{conv}_{3-SAT}$, instances of the random $3$-SAT problem that are not able to find a full convergence of the messages. This property allows us to understand the worst-case scenario of the SP algorithm in this region. Therefore, we analyzed a set of instances, with cardinality $2\, 10^3$, with $N=10^4$, and we looked at the instances where the average error of convergence was not trivial, i.e. different from 0. The plot shows that in the region $\alpha \in [4.200, 4.355)$ the average error of convergence $\langle \epsilon \rangle$ is bounded, i.e. $\langle \epsilon \rangle \leq 0.15$, while beyond the threshold $\alpha^{conv}_{3-SAT}$ the quantity $\langle \epsilon \rangle$ grows linearly with $\alpha$. This fact suggests that some information can also be extracted by those messages, although it is not completely correct. However, if we use the local information obtained by those messages, could we find an assignment of the Boolean variables better than a random one? To answer this question, we create a simple neural network described in the next subsection.

\subsection{The Neural Network and Numerical Analysis}

Algorithms based on the theory of deep learning have become essential in a wide variety of scientific disciplines. Deep learning is a class of machine learning algorithms that uses multiple layers to extract higher-level features from the raw input. In this manuscript, the deep learning methodology extracts patterns for fixing a single Boolean variable by building a function learned by solutions coming from Survey Inspired Decimation Algorithm.  Deep learning theory is based on Artificial Neural Networks (ANN), a series of functional transformations. These functional transformations can be obtained by fixing a set of basis functions in advance and allowing them to be adaptive during training. 

\begin{figure}
 \centering
 \includegraphics[width=0.45\columnwidth]{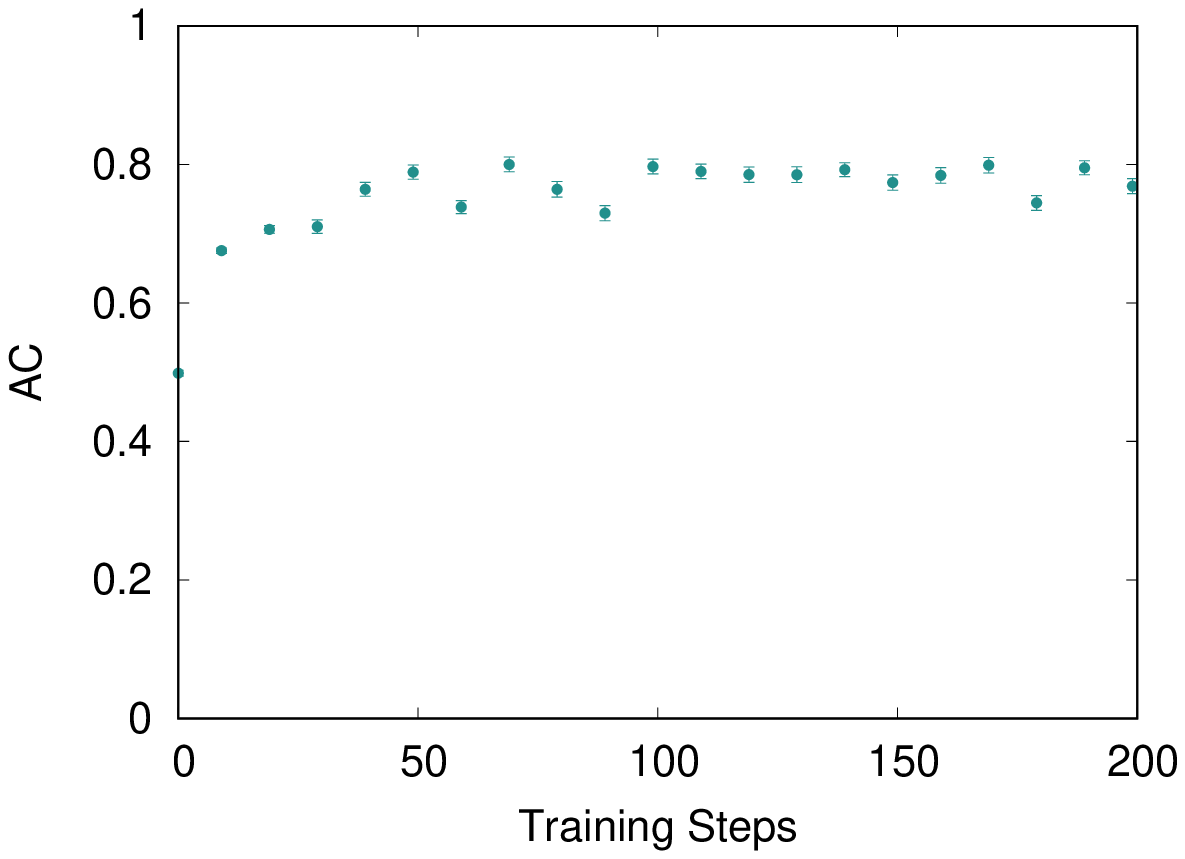}
  \includegraphics[width=0.45\columnwidth]{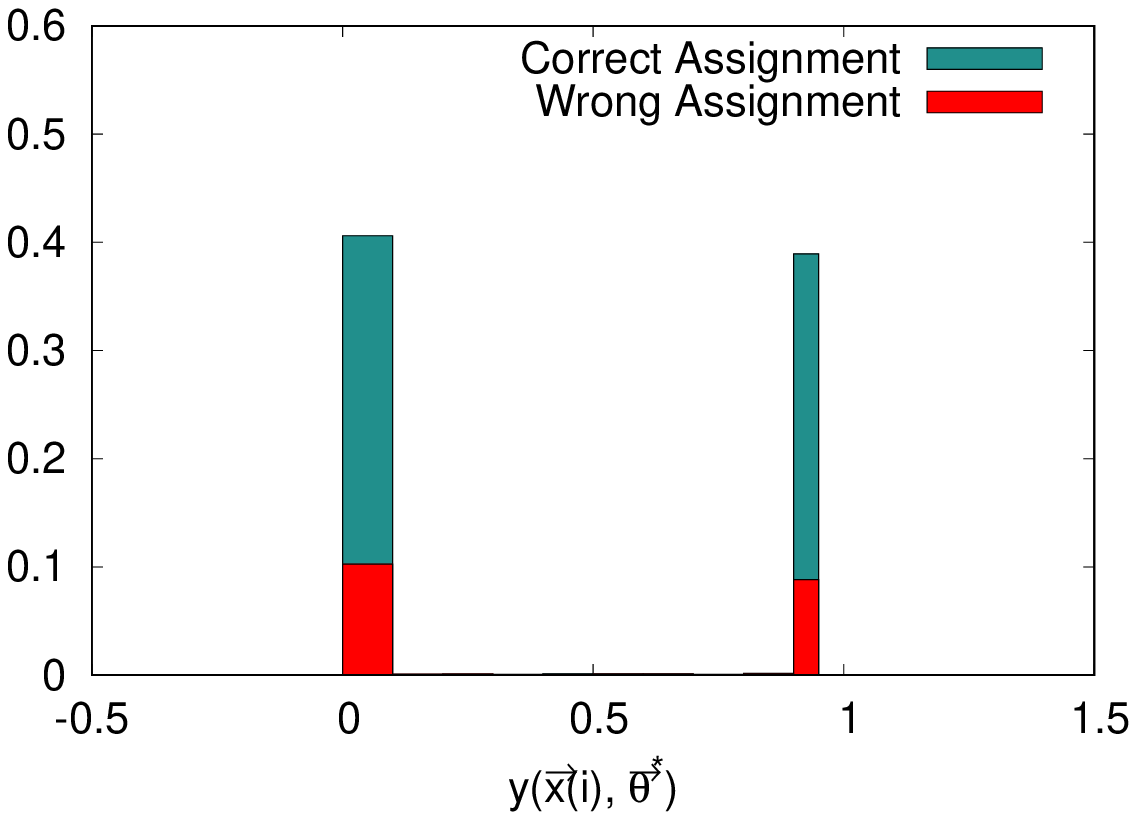}
  \caption{\textbf{Left}: The figure displays the accuracy (AC) as a function of the number of training steps. In this case, for each training step, i.e., for each time that we call the SGD for optimizing the parameter $\vec{\theta}$ on a batch of the training data set, we compute the accuracy  (equation \eref{accuracy}) on a validation/test set. The validation/test set is composed by $36$ instances of the random $3$-SAT with $\alpha=4.23$ and $N=10^4$. As shown by the plot, the maximum accuracy of the neural network is reached just after $100$ training steps. We run the training procedure for one epoch composed of $10^4$ training steps. Error bars are standard deviations. \textbf{Right}: The figure shows the normalized histogram of the conditional probability returned by the neural network $y(\vec{\mathbf{x}}(i), \vec{\theta}^*)$ that a variable $i$ must be set to $TRUE$ or not. In green, we plot the correct assignments, i.e. the value of $y(\vec{\mathbf{x}}(i), \vec{\theta}^*)$ which allows us to correctly fix the variable $i$ to $TRUE$ or $FALSE$, while in red we plot the value of $y(\vec{\mathbf{x}}(i), \vec{\theta}^*)$ which makes us guessing a wrong assignment. Surprisingly, only the $\sim 20\%$ of variables is fixed in the wrong way.}
  \label{fig::AC}
\end{figure}

\begin{figure}
 \centering
 \includegraphics[width=0.45\columnwidth]{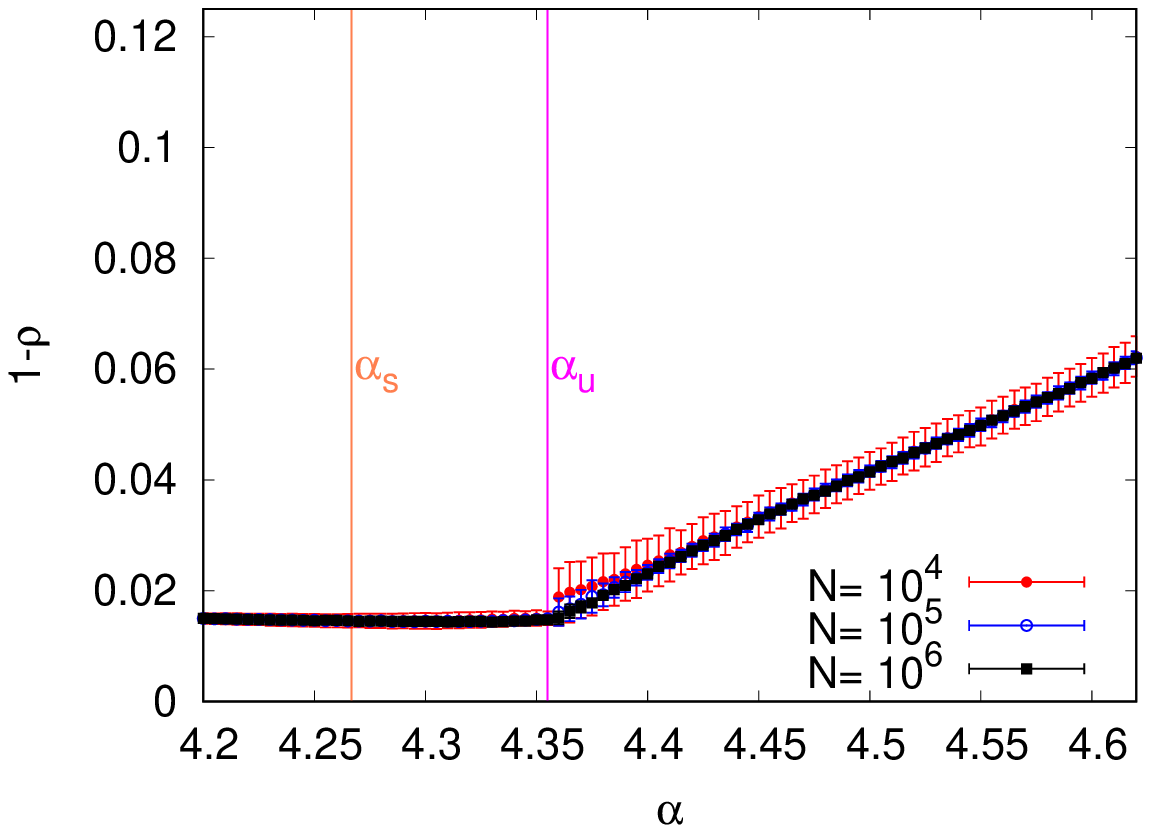}
  \includegraphics[width=0.45\columnwidth]{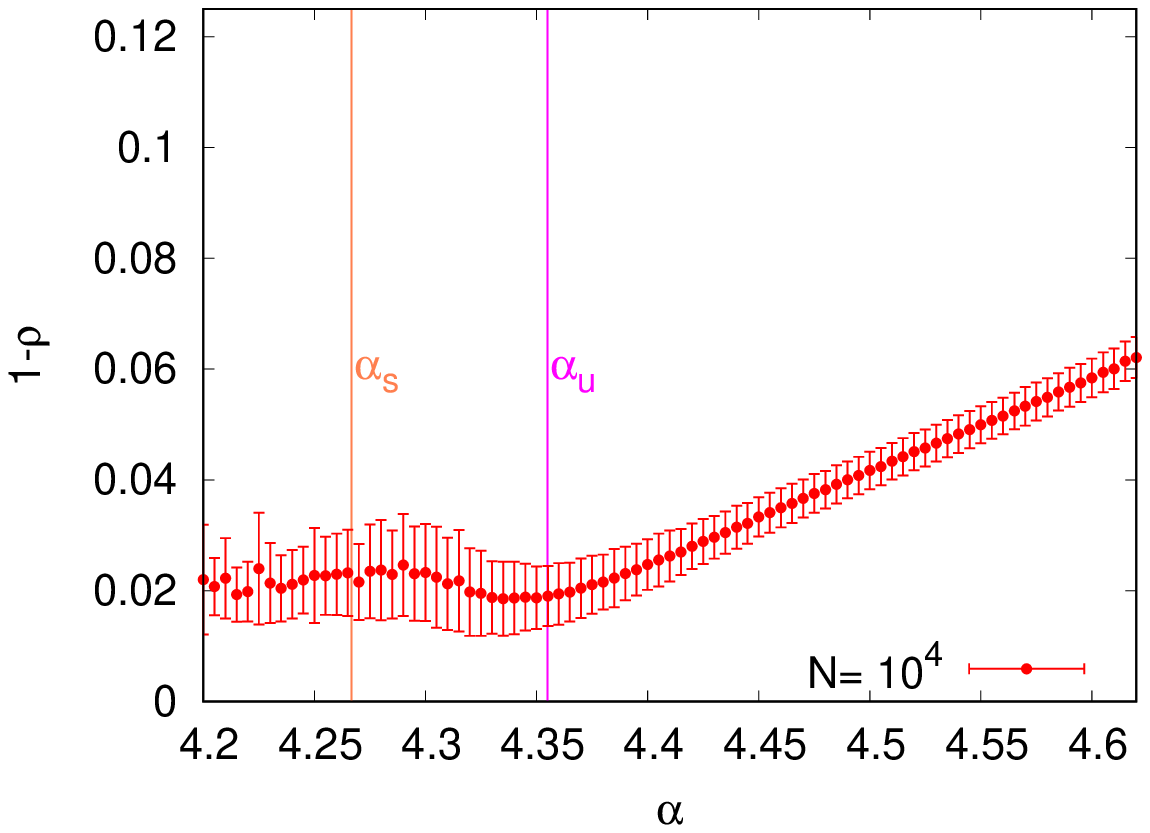}
  \caption{\textbf{Left}: The plot displays the quantity $1-\rho$ as a function of the clause density $\alpha$. For the random MAX-E-$3$-SAT problem, we analyzed $10^3$ instances for $N=10^4$ (red points), $10^2$ instances for $N=10^5$ (blue points), and $10$ instances for $N=10^6$ (black points). The vertical coral line identifies the SAT-UNSAT threshold at $\alpha_s=4.267$. The vertical magenta line identifies the SP unique solution threshold at $\alpha_u=4.355$. Error bars are standard deviations. In the region $\alpha \in [4.200, 4.355)$, the average was performed only on the instances of MAX-E-$3$-SAT problem that converged, i.e., $t^*<t_{max}$. In the region $\alpha \in [4.355, 4.620]$, the average was performed only on the instances of random $3$-SAT problem that did not converge, i.e., $t^*=t_{max}$. \textbf{Right}: The plot displays the quantity $1-\rho$ as a function of the clause density $\alpha$ for instances of the random MAX-E-$3$-SAT problem that did not converge. We analyzed $2\,10^3$ instances for $N=10^4$ (red points), and we averaged only on the instances that after $t^*=t_{max}$ did not find a convergence for all the messages. The vertical coral line identifies the SAT-UNSAT threshold at $\alpha_s=4.267$. The vertical magenta line identifies the SP unique solution threshold at $\alpha_u=4.355$. Error bars are standard deviations. }
  \label{fig_main}
\end{figure}

In our case, the ANN is a feed-forward neural network that is trained as a classifier to predict the conditional probability that a variable $i$ must be set to $TRUE$ or not, given a piece of local information expressed into a vector of input data $\vec{\mathbf{x}}(i)$. This vector $\vec{\mathbf{x}}(i)$ has $4$ dimensions and it is composed by $\vec{\mathbf{x}}(i)=[1-\pi_i^{+},1-\pi_i^{-},n_i^{+},n_i^{-}]^{T}$. $1-\pi_i^{\mp}$, under the assumption that the factor graph is locally tree-like, may be interpreted as the probability that the variable $i$ does not receive warnings from the set of clauses where it appears negated ($n^{-}$) or not negated ($n^{+}$). The components of the vector $\vec{\mathbf{x}}(i)$ could be interpreted as the features used for feeding a deep neural network in the general framework of machine learning.

The deep neural network $\mathcal{NN}(\vec{\mathbf{x}}(i), \vec{\theta})$ has five layers. The input and output layers have 4 and 1 neuron respectively, where sigmoidal activation function acts element-wise on each neuron, i.e. $\sigma(a)=(1+\exp(-a))^{-1}$. The hidden layers are sigmoidal layers composed by 40 neurons each. The total set of parameters to be optimized is defined with $\vec{\theta}$. We define as loss function the following cross-entropy error function:
\begin{equation}
\label{eq::errorfunc}
E(\vec{\theta})=-\sum_{i=1}^{\mathcal{N}} \{ \mathcal{T}_i \ln y_i(\vec{\mathbf{x}}(i), \vec{\theta}) + (1 - \mathcal{T}_i) \ln (1-y_i(\vec{\mathbf{x}}(i), \vec{\theta})) \},
\end{equation}
where $\mathcal{T}$ is the target variable, and $\mathcal{N}$ is the total batch size. In our case, the target variables $\mathcal{T}_i$ are the variables into a satisfiable assignment of a random $3$-SAT problem. We chose the cross-entropy error function as a loss function because we are dealing with a classification problem. Classification is the problem of identifying to which of a set of categories a new observation belongs, based on a training set of data containing observations whose category membership is known. In our case, the training set is composed by vectors $\vec{\mathbf{x}}(i)$ and targets that are Boolean variables. We can assume that each of these targets has a Bernoulli distribution. Considering them composed by independent observations, the loss function that arises naturally by taking the negative log-likelihood is the cross-entropy error function.

The output $y(\vec{\mathbf{x}}(i), \vec{\theta})$ can be interpreted as the conditional probability $p(x_i=1|\vec{\mathbf{x}}(i))$, with $p(x_i=0|\vec{\mathbf{x}}(i))$ given by $1-y(\vec{\mathbf{x}}(i), \vec{\theta}) $, where $\vec{\theta}$ is the set of parameters that has to be optimized.

For optimizing these parameters, we need to train our neural network. For doing that, we solved $400$ random $3$-SAT problems at $\alpha=4.2$ and $N=10^4$ using SID (see \Sref{sec::SPalgo}). For each of this $400$ instances we stored one solution and the $\vec{\mathbf{x}}(i)$ for each variable $i$. In other words, for each instance random $3$-SAT we have $10^4$ vectors $\vec{\mathbf{x}}(i)$, and at each of these vectors is associated the target variable $\mathcal{T}_i$, which is the Boolean variable associated with the satisfiable assignment. The training of the neural network was performed by giving a batch of $20$ random vectors $\vec{\mathbf{x}}(i)$ and the respective target variables $\mathcal{T}_i$, without replacement, to the neural network. The optimal assignment of $\vec{\theta}$, i.e., $\vec{\theta}^*$, to which the right-hand side of equation \eref{eq::errorfunc} vanishes, can be found by running an SGD algorithm. 

For our simulations, we used the default SGD (Adam \cite{kingma2014adam}) given by mlpack \cite{mlpackpaper}. For testing the performance of the deep neural network, we calculated the accuracy in computing the conditional probability $p(x_i=1|\vec{\mathbf{x}}(i))$ on a validation data set. The validation data set was obtained by solving $36$ random $3$-SAT problems at $\alpha=4.23$ and $N=10^4$ using SID. We define the accuracy of the neural network as:
\begin{equation}
\label{accuracy}
AC=\frac{1}{\mathcal{M}}\sum_{s=1}^{\mathcal{M}}\mathcal{NH}(\mathcal{S}_s,\mathcal{Y}_s),
\end{equation}
where, $\mathcal{M}$ is the total number of test solutions, i.e., the $36$ solutions of random $3$-SAT problems at $\alpha=4.23$ and $N=10^4$; $\mathcal{NH}(a,b)$ is the normalized Hamming distance between two strings $a$ and $b$ with the same length; $\mathcal{S}_s$ is the exact solution obtained by SID; $\mathcal{Y}_s$ is the approximate solution obtained from the deep neural network. As the reader can see, we tested the deep neural network on solutions with a different value of $\alpha$ to which the neural network was trained. It was possible because we assumed that the local information obtained by SP equations should be independent of the clause density $\alpha$.

In the left panel of \Fref{fig::AC}, we present the accuracy (AC) of the neural network on the validation data set of $36\, 10^4$ elements as a function of the number of training steps, i.e., the number of times that we called the SGD for optimizing the set of parameters $\vec{\theta}$. The neural network starts by giving a random assignment to the Boolean variables, AC $\sim 50\%$, and after $100$ training steps, it learns how to assign the Boolean variables. We also tested the accuracy of the neural network on a validation data set of $17\, 10^4$ elements coming from $17$ solutions of $3$-SAT at $\alpha=4.24$ and $N=10^4$, obtaining the same accuracy. However, as shown on the right panel of \Fref{fig::AC}, the approximation of the conditional probability $y(\vec{\mathbf{x}}(i), \vec{\theta}^*)$ completely fails for $\sim 20\%$ of the variables. This failure is not bad for our purpose. Indeed, we are not interested in building a solution that satisfies all the clauses. Still, we are interested in finding an approximation of a solution that minimizes the number of unsatisfied clauses into an instance of MAX-E-$3$-SAT.

The whole algorithm, which we name \textbf{DeepSP}, is presented in Algorithm~\ref{algoDeepSP}.

 \begin{minipage}{0.85\textwidth}
 \begin{algorithm}[H]
 \label{algoDeepSP}
\SetAlgoLined
\KwIn{A CNF formula for MAX-E-$3$-SAT.}
\KwOut{An assignment Sol$_{MAX-E-3-SAT}$ for MAX-E-$3$-SAT.}
Learn the parameters $\vec{\theta}^*$ of the neural network $\mathcal{NN}(\vec{\mathbf{x}}(i), \vec{\theta})$\;
Random inizialization of all messages $\eta_{a \to i}$ in the clauses.\;
 \For{$t<{t_{max}}$}{
 RUN SP on the factor graph underlying the CNF formula, i.e. Algorithm~\ref{algoSP}\;
 }
 \For{$1\leq i \leq N $}{
 Compute $\vec{\mathbf{x}}(i)$\;
 Use the output of $\mathcal{NN}(\vec{\mathbf{x}}(i), \vec{\theta}^*)$ for fixing the variable $i$, i.e. if $y(\vec{\mathbf{x}}(i), \vec{\theta})\geq 0.5$ then $i$ is set to $1$, else $i$ is set to $0$\;
 Save the value of $i$ into Sol$_{MAX-E-3-SAT}$ \;
 } 
 \Return Sol$_{MAX-E-3-SAT}$ \;
 \caption{\textbf{DeepSP}}
\end{algorithm}
 \end{minipage}

For speeding up the algorithm, we introduced the convergence criterion explained in \Sref{sec::SPalgo}.
The computational complexity of \textbf{DeepSP} algorithm is, therefore, $\Theta(N)$. Indeed, the maximum number of operation that it takes for outputting a result, after the training procedure for optimizing the parameters $\vec{\theta}$ of the deep neural network, is $\sim t_{max}kM + \Or(|\vec{\theta}|^2)N$, where $|\vec{\theta}|<<N$ and $M=\alpha N$. Moreover, once the deep neural network's training procedure is performed, one can save the parameters' value and upload them instead of re-training the neural network each time. We also release the parameters' value, which can be downloaded from \cite{Marino2019GITHUB}.

For performing an analysis on the performance of this heuristic-learning algorithm, we need, therefore, to check the ratio of the number of satisfied clauses to the total number of clauses, i.e., $\rho$. This result is described in \Fref{fig_main}. Both plots describe the quantity of $1-\rho$ as a function of $\alpha$. In each plot, having $1-\rho$ on the $y$-axis, the top end of the plot coincides with the estimate of the random assignment threshold, i.e., $1-\rho_{rand}=1-7/8=0.125$, to provide an immediate indication of the performance of the \textbf{DeepSP}. In the left panel of the \Fref{fig_main}, in the region $\alpha \in [4.200, 4.355)$ the average of $1-\rho$ is performed only on the instances of random $3$-SAT problem that converged, i.e. $t^*<t_{max}$. In other words, we are looking at the average-case performance of the heuristic-learning algorithm.

The behavior of $1-\rho$ is constant, showing, therefore, that the \textbf{DeepSP} algorithm can find approximate solutions such that only $1.46\%$ of clauses are unsatisfied by the assignment found, more precisely $(1-\rho)=0.0146 \pm 0.0002$. 
In the region $\alpha \in [4.355, 4.620]$ the average was performed only on the instances of random $3$-SAT problem that did not converge, i.e., $t^*=t_{max}$. This is obvious because no convergence is possible beyond $\alpha^{conv}_{3-SAT}$. The behavior of $1-\rho$, in this case, is not constant anymore, but it is linear with the clause density $\alpha$. 

In the right panel, we plot, instead, the worst-case scenario for the SP equations, i.e., only the instances that did not find a convergence of all messages, i.e., the time $t^*=t_{max}$ and $\langle \epsilon \rangle \neq 0$. For showing the worst-case scenario, we run $2\,10^3$ the heuristic-learning algorithm on instances random $3$-SAT with $N=10^4$, and we analyzed only those where a full convergence of the messages was absent. We observe that  the local nature of the SP algorithm helps us to find an approximate solution much better than the one outputted by the Johnson algorithm \cite{johnson1974approximation}. In the region $\alpha \in [4.200, 4.355)$ the average error of the convergence, i.e $\langle \epsilon \rangle$, in the right panel of \Fref{fig::mess_conv} is bounded and the \textbf{DeepSP} seems to follow the same behavior. This behavior shows that the solutions we found using \textbf{DeepSP} are not affected by the loss of convergence. In contrast, in the region $\alpha \in [4.360, 4.620]$ the algorithm, following the behavior of the average error of convergence $\langle \epsilon \rangle$ defined in \eref{averageerror}, is affected to the linear growth, and, therefore, the behavior of the quantity $1-\rho$ grows linearly with $\alpha$. To be more qualitatively, we computed the sample Pearson correlation between two sets of variables:
\begin{equation}
\label{Pearsoncorr}
r_{corr}=\frac{\sum_{i=1}^{n}(x_i-\mu_x)(y_i-\mu_y)}{\sqrt{\sum_{i=1}^{n}(x_i-\mu_x)^2 \sum_{i=1}^{n}(y_i-\mu_y)^2}},
\end{equation}
where $n$ is the sample size, $x_i,\, y_i$ are the individual sample points indexed with $i$, and $\mu_x=\frac{1}{n}\sum_{i=1}^{n}x_i$ the sample mean (and analogously for $\mu_y$), 
between the set of data $1-\rho$ and $\langle \epsilon \rangle$. The sample Pearson correlation is equal to $r_{corr}=0.9959$, confirming that the two quantities are dependent on each other.  

As stated in \Sref{sec::intro}, \textbf{DeepSP} is not competitive with state-of-the-art of MAX-SAT solvers. For showing this, we compared our results with the results obtained by two different established methods: MaxWalkSat \cite{selman1993local}, and SP-$y$ \cite{battaglia2004minimizing}. MaxWalkSat searches for a solution by performing a biased random walk in the solution space. At the same time, SP-$y$ is a message passing algorithm that tries to build a solution assigning variables according to some estimated marginals. 

We start with comparing \textbf{DeepSP} and MaxWalkSat, by analyzing the performance of the two algorithms for instances of MAX-E-$3$-SAT composed by $N=10^6$ variables. In \Tref{Maxwsat}, we show the values of $1-\rho$, the fraction of unsatisfied clauses, at three different values of $\alpha$, i.e. $4.2$, $4.3$, $4.5$. These three points are in three distinct regions of the solution space. The first one, i.e., $\alpha=4.2$, is in the region where solutions always exist. The second one, i.e., $\alpha=4.3$, is in the region where no solution exists, but SP equations always converge. Instead, the third one is in the region where no solution exists, and SP equations do not converge. 

The fraction of unsatisfied clauses obtained by MaxWalkSat is strongly dependent on the cutoff parameter chosen. The \textit{cutoff} parameter in MaxWalkSat identifies the number of flips performed by the algorithm for leading to the greatest decrease in the total number of unsatisfied clauses.  
We chose the values $10^4$, $10^5$, $10^6$, $10^7$, $10^8$ for our comparison. When the cutoff is smaller than or equal to $10^7$, the performance of \textbf{DeepSP} is better than the performance of MaxWalkSat in satisfying the maximum number of clauses. However, from a cutoff of $10^8$, or bigger, the MaxWalkSat performance is the best. 

When the cutoff increases, the time for searching for an optimal solution also increases. This situation makes the MaxWalkSat really hard to analyze analytically. In other words, we do not know when $N\to\infty$ how big the cutoff should be. It is just a parameter that is chosen a priori. In contrast, \textbf{DeepSP} overcomes this issue. No parameter increases its efficiency in finding an optimal approximate solution.   

\begin{table}
\caption{\label{Maxwsat} MaxWalkSat results at $N=10^6$ for different values of $\alpha$ and different cutoffs. We stop increasing the cutoff as soon as we find that MaxWalkSat results are better than the one found by \textbf{DeepSP}.}
\begin{indented}
\item[]\begin{tabular}{ |c|c|c|c|c|} 
\br
cutoff & $N$ &  $(1-\rho)_{\alpha=4.2} \pm \sigma_{(1-\rho)_{\alpha=4.2}}$ & $(1-\rho)_{\alpha=4.3} \pm \sigma_{(1-\rho)_{\alpha=4.3}}$ & $(1-\rho)_{\alpha=4.5} \pm \sigma_{(1-\rho)_{\alpha=4.5}}$ \\
\mr
$10^4$ &$10^6$& 0.5074 (6) & 0.5197 (7) & 0.5456 (8) \\ 
$10^5$ &$10^6$& 0.3760 (6)&0.3880 (6) &0.4137 (6) \\ 
$10^6$ &$10^6$& 0.0872 (3) &0.0971 (3) &0.1176 (4) \\ 
$10^7$ &$10^6$& 0.0240 (3) &0.0320 (4) &0.0497 (3) \\
$10^8$ &$10^6$& 0.0062 (1) &0.0137 (3) &0.0340 (4) \\ 
\br
\end{tabular}
\end{indented}
\end{table}

SP-$y$ is a message passing algorithm ideated for minimizing the number of violated clauses in the MAX-E-$k$-SAT. It takes as INPUT a Boolean formula $\mathcal{F}$ in conjunctive normal form and outputs a simplified Boolean formula $\mathcal{F'}$ in conjunctive normal form and a partial truth-value assignment for the variables. If $\mathcal{F'} \neq \emptyset$ is given to a heuristic MAX-SAT Solver (as MaxWalkSat, Simulated Annealing) for building the complete assignment of $\mathcal{F}$. It uses decimation and backtracking strategies, which means that iteratively fixes and un-fixes variables for building up a partial (or complete) solution of $\mathcal{F}$ according to some estimated marginals.

Its performance is extraordinary. For $\alpha=4.24$ and $N=10^5$, SP-$y$ can find a completely satisfiable assignment of the formula $\mathcal{F}$, or, in its worst performance, a value of $(1-\rho)\approx 10^{-5}$. Above the $SAT-UNSAT$ threshold, for example, when $\alpha=4.29$, the best performance of the algorithm, on a single sample of a MAX-E-$3$-SAT instance, reached a value of $(1-\rho)\approx 2\,10^{-4}$. These performances can be obtained only when backtracking and decimation moves are performed, implying a very long run-time. Indeed, the SP-$y$ algorithm has a computational complexity of order $O(N^2)$, making the established method unfeasible for huge values of $N$.  

Without any decimation or backtracking moves, as in \textbf{DeepSP}, the algorithm has the same performances of the heuristic MAX-SAT Solver used (indeed $\mathcal{F}=\mathcal{F'}$ ), and the performance of the algorithm is strongly dependent on the parameters of the heuristic. 

\begin{figure}
 \centering
 \includegraphics[width=0.45\columnwidth]{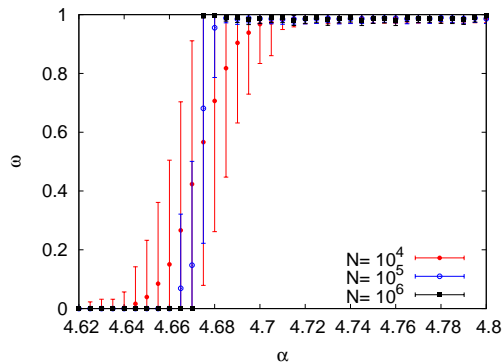}
  \caption{The plot shows the behavior of $\omega$, i.e., the fraction of variables where unit propagation strategy was used for fixing the value of the Boolean variable $i$ in a MAX-E-$3$-SAT instance, as a function of the clause density $\alpha$. For the random MAX-E-$3$-SAT problem, we analyzed $10^3$ instances for $N=10^4$ (red points), $10^2$ instances for $N=10^5$ (blue points), and $10$ instances for $N=10^6$ (black points). Error bars are standard deviations.}
  \label{fig_up}
\end{figure}

For concluding the analysis of the algorithm, we looked at the performance of the algorithm beyond $\alpha=4.620$. In the region $\alpha \in (4.620, +\infty)$, we meet a point at $\alpha=4.67$, for $N\to \infty$, where the SP equations return at least one message $\eta_{a \to i}=1$, also if $n_i > 1$. When such an issue happens, numerical instability into the SP equations appears. For avoiding this issue, therefore, we introduce a simple strategy, i.e. \textit{unit propagation} strategy, on the variable $i$ where at least one message $\eta_{a \to i}=1$ appears in the set of functional nodes associated with it, and we fix such a variable by using the rule: $i$ is $TRUE$ if $n_{i}^{+}> n_{i}^{-}$, $FALSE$ otherwise. In \Fref{fig_up} we present the fraction of variables where unit propagation strategy was used for fixing the value of the Boolean variable $i$ in a sample of MAX-E-$3$-SAT instances, i.e., $\omega$, as a function of $\alpha$. At $\alpha=4.67$, for $N=10^6$, we pass from a region where \textbf{DeepSP} uses the neural network to a region where the unit propagation rule fixes all the variables. Beyond the threshold at $\alpha=4.67$ we can claim, also if numerically we meet the random assignment threshold $\rho_{rand}=7/8$ at $\alpha=10$ for each value of $N$ analyzed, that the worst-case performance of the \textbf{DeepSP} is equal to the random assignment outputted by the Johnson algorithm \cite{johnson1974approximation}.

\section{Conclusion}

This paper has presented a new heuristic-learning algorithm, namely \textbf{DeepSP} algorithm, that finds approximate solutions for the MAX-E-$3$-SAT problem. This algorithm runs SP equations on the random factor  graph associated with the MAX-E-$3$-SAT problem. It gives the local information computed to a neural network $\mathcal{NN}(\vec{x}(i), \vec{\theta}^*)$. The set of parameters $\vec{\theta}$ are optimized following a supervised learning approach ( by using target values obtained by SID on a sample of random $3$-SAT problems) and outputs an assignment. We have displayed an accurate analysis to explain the algorithm's average and worst-case behavior as a function of the clause density $\alpha$. We have started with presenting the limits of the SP equations and the neural network's performance in learning and inferring the conditional probability that a variable $i$ should take to $TRUE$ or $FALSE$. Then, we have shown that this algorithm can find approximate solutions that outperform the random assignment threshold value in the region where the SP equations do not present any numerical instability, and we have identified its algorithmic threshold, which is the ultimate limit of the algorithm, at $\alpha_a=4.67$. Moreover, we have observed that the algorithm's output is strongly related to the average error of convergence $\langle \epsilon \rangle$ that the SP equations commit if they do not find a unique set of fixed points. Although this algorithm is not competitive with state-of-the-art MAX-SAT solvers, it can solve substantially larger and more difficult problems than it ever saw during training.

As future research directions, we propose to analyze the performance of the algorithm on MAX-E-$4$-SAT problem and verify if a Belief Propagation algorithm version could perform as well as our \textbf{DeepSP} on that particular problem. We also suggest using SP-$y$ equations, instead of SP equations, for improving the performance of this new heuristic-learning algorithm. Moreover, we suggest analyzing with a probabilistic approach the properties of the maximum error of convergence $\epsilon_{max}$ that the SP equations can perform on random instances of the MAX-E-$k$-SAT problem. This maximum error should be related to the algorithmic performance, as numerically shown in our analysis.

\ack{R. M. acknowledges interesting discussions with Nicolas Macris.
This work is supported by the Swiss National Foundation grant number 200021E 17554.}

\section*{Data Availability Policy}
The data that support the findings of this study are available from the corresponding author upon reasonable request. The code that supports the findings of this study is openly available at \cite{Marino2019GITHUB}.

\section*{References}
\bibliographystyle{iopart-num}
\bibliography{references}

\end{document}